\documentclass{amsart}

\usepackage{graphicx}
\usepackage{textcomp}
\usepackage{xcolor}
\usepackage{cite}
\usepackage[foot]{amsaddr}

\usepackage[pagebackref,breaklinks,colorlinks]{hyperref}

\usepackage[capitalize]{cleveref}
\crefname{section}{Sec.}{Secs.}
\Crefname{section}{Section}{Sections}
\Crefname{table}{Table}{Tables}
\crefname{table}{Tab.}{Tabs.}

\title{A Vulnerability of Attribution Methods \\ Using Pre-Softmax Scores}

\author{Miguel Lerma$^1$}
\address{$^1$Northwestern University, Evanston, USA}
\email{$^1$mlerma@math.northwestern.edu}
\author{Mirtha Lucas$^2$}
\address{$^2$DePaul University, Chicago, USA}
\email{$^2$mlucas3@depaul.edu}

\date{\today}

\begin{document}

\begin{abstract}
We discuss a vulnerability involving a category of attribution methods used to
provide explanations for the outputs of convolutional neural networks 
working as classifiers. 
It is known that this type of networks are vulnerable to adversarial attacks, 
in which imperceptible perturbations of the input
may alter the outputs of the model.  In contrast,
here we focus 
on effects that small modifications in the model may cause on
the attribution method without altering the model outputs.
\end{abstract}

\maketitle

\section{Introduction}

The black box nature of current artificial intelligence (AI) models
is considered problematic in areas with low tolerance to errors, such as
Computer Aided Diagnosis (CAD) and autonomous vehicles.  To palliate the
effect of mistakes and increase confidence in the model, explanation methods
have been developed to justify the model outputs \cite{burkart2021}.

\begin{figure*}[htb]
\centering
\ \includegraphics[height=1.5in]{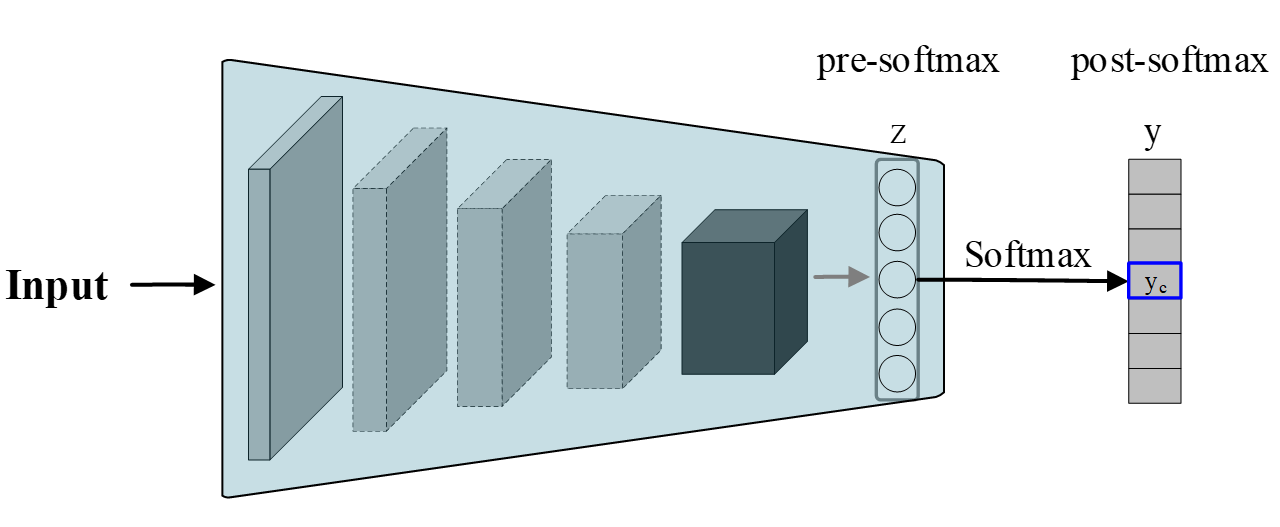}
\caption{Structure of a typical classifier network. After a number of
  convolutional blocks this kind of network ends with a fully
  connected network producing a (pre-softmax) output \textbf{z},
  followed by a softmax activation function with (post-softmax) output
  \textbf{y}.}\label{f:dnn1}
\end{figure*}

A class of explanation methods widely used on convolutional neural networks (CNN)
take the form of attribution methods
that determine how much different parts of the input of a model contribute
to produce its final output. In general, the networks on which 
these methods are used consist of several
convolutional layers that produce a vector of outputs $\mathbf{z} = (z_1,z_2,\dots,z_n)$,
which is then transformed with a softmax function into a vector of probabilities 
$\mathbf{y} = (y_1,y_2,\dots,y_n)$, where $n$ is the number of classes.
(Figure\,\ref{f:dnn1}). 
Each post-softmax output can be interpreted as the amount of confidence
about the input sample belonging to each of the several classes $1,2,\dots,n$.
In classification tasks, the output with maximum value corresponds to the
class to which the input sample is considered to belong.

Gradient-based attribution
methods for convolutional networks work by computing the gradient
$\nabla_{\mathbf{x}}S = (\partial S/\partial x_1,\dots,\partial
S/\partial x_N)$ of an output or ``score'' $S$ of the network respect
to a set of inputs or unit activations $\mathbf{x} = (x_1,\dots,x_N)$,
where $N$ is the number of inputs or internal units, and 
$S$ may represent either one of the pre-softmax outputs $z_i$,
or one of the post-softmax outputs $y_i$.
The assumption is that each derivative $\partial S/\partial x_i$
provides a measure of the impact of $x_i$ on the score~$S$.
A few examples of attribution methods using this approach are Grad-CAM \cite{selvaraju2019}, 
Integrated Gradients (IG) \cite{sundararajan2017},
and RSI~Grad-CAM \cite{lucas2022}.

In \cite{lerma2023} there is a detailed analysis of the differences between using
gradients of pre-softmax versus post-softmax outputs.  In that paper it is argued
that the post-softmax version of gradient-based methods is more robust 
and not affected by a vulnerability suffered by the pre-softmax version.
Here we will provide a brief overview of the main argument leading to that conclusion, and a specific way in which the vulnerability could be exploited.

\section{Previous Work}

The possibility of fooling a classification network with
\emph{adversarial attacks} by using slightly modified inputs
is well known \cite{goodfellow2015, akhtar2018}.  On the other hand,
the ability of altering the output of an attribution method without modifying
the model predictions has not been studied in the same extent, but there
are also some findings in that direction (see e.g. \cite{kindermans2022,srinivas2021rethinking}). 
Since terminology
may vary across works we must clarify that we use the term \emph{attribution}
method where other authors use \emph{explanation} or \emph{interpretation}
method. We made this decision to stress the fact that an attribution method
may not quite fulfill human expectations for an explanation, in particular
Grad-CAM-like methods seem to do a good job in locating the parts of 
an input containing a sample of a class, i.e., it helps to determine 
\emph{where} the object
corresponding to the class predicted by the model is in the input image,
but that does not necessarily explains \emph{why} the output of the
network is what it is.  However, when citing a work we keep the authors
terminology in this regard.

In \cite{ghorbani2019} adversarial attacks against interpretation methods
are tried and tested. They work in a similar way to adversarial attacks
against network predictions, the main idea is to search for small perturbations
of sample inputs that change the output of interpretation methods without altering
the network predictions. The work is mainly experimental and requires extensive
testing.

The works mentioned above focus on how perturbation of inputs can 
alter outputs of attribution methods.  On the other hand,
the authors of \cite{heo2019} study the possibility of fooling
interpretation methods by adversarial model manipulation
without perturbing model accuracy.
Their approach consists of applying fine tuning to a given
model with a loss term that includes the interpretation results
in the penalty term of the objective function.  So, rather than 
perturbing inputs the approach of the authors is to perturb the model itself.
Again, the work is mainly empirical and requires extensive testing.

In \cite{kindermans2022} it is shown that adding a constant shift 
to the input data has no effect on the model but causes numerous 
attribution methods to produce incorrect attributions.

Concerned with the quality of explanation methods, the authors of
\cite{hedstrom2023} have built \texttt{Quantus}, a comprehensible tool
for XAI evaluation, and they list a number of metrics that can be applied 
to explanation methods.  The metric that is most closely related to our work
is \emph{robustness}, which (in their words) 
\emph{measures to what extent explanations
are stable when subject to slight perturbations in the input,
assuming that the model output approximately stayed the same.}
As indicated, the metric is based on the effects of perturbations
applied to input samples.

Before showing the details of our work we state how it differs
from previous work in identifying possible adversarial attacks against
attribution methods.  First, our work does not require to perturb inputs.
Second, our method does not require training or fine tuning a model.
We just identify a vulnerability of Grad-CAM-like methods using 
pre-softmax scores, and show how the model can be modified to exploit the
vulnerability. Going beyond the theory we show an specific modification
that has the desired effect, and illustrate it with several examples
as a proof of concept.

\section{A vulnerability of attribution methods using pre-softmax scores.}

In this section we examine a vulnerability that affects attribution methods for CNNs that 
work with pre-softmax scores, with a special emphasis on gradient-based methods, although many of
the considerations can be easily extended to methods that work with finite differences rather
than gradients, such as Layer-wise Relevance Propagation (LRP) \cite{montavon2019} 
and DeepLIFT \cite{shrikumar2017}.

\begin{figure*}[htb]
\centering
\ \includegraphics[height=1.5in]{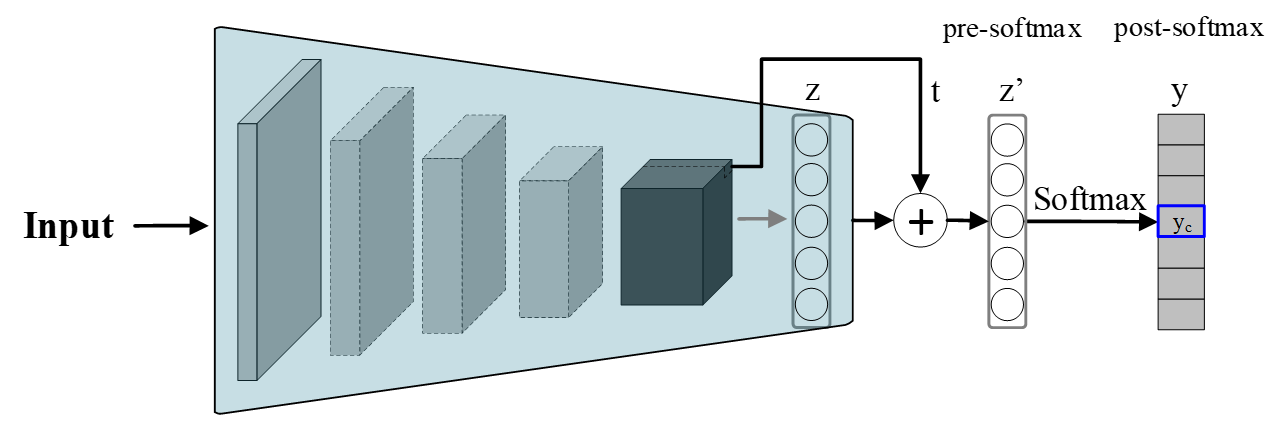}
\caption{Example of alteration of a classifier network that changes attributions based on pre-softmax scores
without changing post-softmax scores.}\label{f:dnn2}
\end{figure*}

\subsection{The softmax function}

The output of the softmax function applied to a vector $\mathbf{z} = (z_1,z_2,\dots,z_n)$
is the vector $\mathbf{y} = (y_1,y_2,\dots,y_n)$ whose components are:
\begin{equation}\label{e:softmaxdef}
  y_c = \frac{e^{z_c}}{\sum_{i=1}^n e^{z_i}}
  \,.
\end{equation}
The outputs of the softmax verify $0< y_c < 1$ for all classes $c=1,\dots,n$,
and $\sum_{c=1}^n y_c = 1$, so the
$y_c$ are usually interpreted as probabilities.

Note that adding an amount $t$ independent of the class $i$ to all the arguments
of the softmax, $z'_i = z_i + t$, has no effect on its outputs:
\begin{equation}\label{e:L4}
\begin{aligned}
  y'_c &= \frac{e^{z'_c}}{\sum_{i=1}^n e^{z'_i}} 
  = \frac{e^{z_c+t}}{\sum_{i=1}^n e^{z_i+t}} 
  = \frac{e^t \, e^{z_c}}{\sum_{i=1}^n e^t e^{z_i}} \\
  &= \frac{e^t \, e^{z_c}}{e^t \sum_{i=1}^n e^{z_i}}
  = \frac{e^{z_c}}{\sum_{i=1}^n e^{z_i}} = y_c
  \,.
\end{aligned}
\end{equation}
So, the change $z_i \mapsto z_i+t$ for every $i$ does not change the
network post-softmax outputs $y_c$.  Note that $t$ does not need to be a constant, all that is required is that $t$ is independent of $i$.

Since adding $t$ has no effect in the output of the softmax,
the derivatives of
the outputs of the softmax won't change after adding $t$ to its arguments:
\begin{equation}
\frac{\partial y'_i}{\partial x} = \frac{\partial y_i}{\partial x} 
\,,
\end{equation}
however the derivatives of the pre-softmax $z_i$ may change:
\begin{equation}
\frac{\partial z'_i}{\partial x} = \frac{\partial (z'_i + t)}{\partial x} 
= \frac{\partial z_i}{\partial x} + \frac{\partial t}{\partial x} 
\,,
\end{equation}
so that $\frac{\partial z'_i}{\partial x} \neq \frac{\partial z_i}{\partial x}$
if $\frac{\partial t}{\partial x} \neq 0$.

This theoretical result and its potential impact
in gradient-based attribution methods
are carefully examined in~\cite{lerma2023}, and
it is also mentioned in \cite[sec.\,2]{srinivas2021rethinking}).
In the following section we will 
provide a proof of concept showing
how this results can be used to radically modify a heatmap
produced by an attribution method such as Grad-CAM.

\begin{figure*}[htb]
\centering
\ \includegraphics[width=4in]{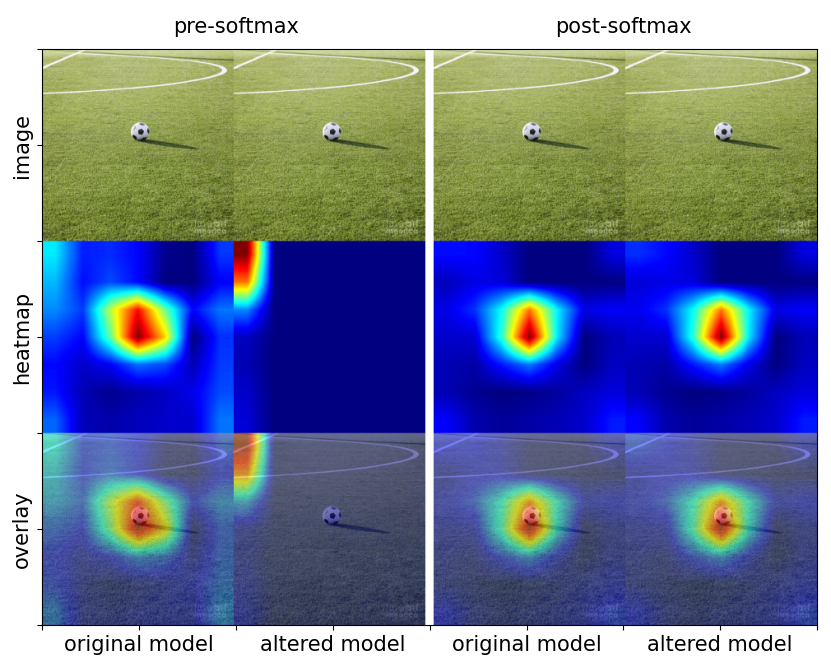}
\caption{Heatmaps produced by Grad-CAM using pre-softmax and
post-softmax outputs respectively, intended to locate 
the position of the soccer ball. The original model is a VGG19 network
pretrained on ImageNet. The altered model is the same VGG19 network
slightly modified, but still functionally equivalent (same final outputs) to
the original network. The heatmaps are computed at the last convolutional
layer of each model. Note that Grad-CAM working on pre-softmax outputs has been
tricked to produce wrong heatmaps.  The heatmaps obtained using post-softmax outputs
remain unchanged.}\label{f:soccer}
\end{figure*}

\subsection{A vulnerability of attribution methods using pre-softmax scores.}

Equation (\ref{e:L4}) shows that the softmax function has no unique
inverse because we can add to its arguments $z_1,\dots,z_n$ any scalar $t$ independent of $i$ 
without changing the output of the softmax.

In the example shown here (\figurename\,\ref{f:dnn2}) 
the network is a VGG19 pretrained on ImageNet \cite{simonyan2015}.
Then, $t$ is the result of adding 
the activations of the units placed in position $(0,0)$ of the final pool layer
(block5\_pool) across all its channels multiplied by a constant $K$.
More specifically, if $A_{ijk}$ presents the activation of unit in position $(i,j)$ of channel $k$
of the last pooling layer, then:
\begin{equation}
t = K \sum_{k} A_{00k}
\,,
\end{equation}
where $K$ is a constant---in our experiment we used $K=10$.

After $t$ is added to the original
$z_i$ pre-softmax scores of the network we get new pre-softmax scores $z'_i = z_i+t$.  
This makes the new pre-softmax scores strongly dependent on the units 
in position $(0,0)$ of the final pool layer without altering the post-softmax scores of the network.
Consequently, we expect that heatmaps produced by Grad-CAM to strongly highlight the upper left area of
the image regardless of whether that part of the image is related to the network final output.

\begin{figure*}[htb]
\centering
\ \includegraphics[width=4in]{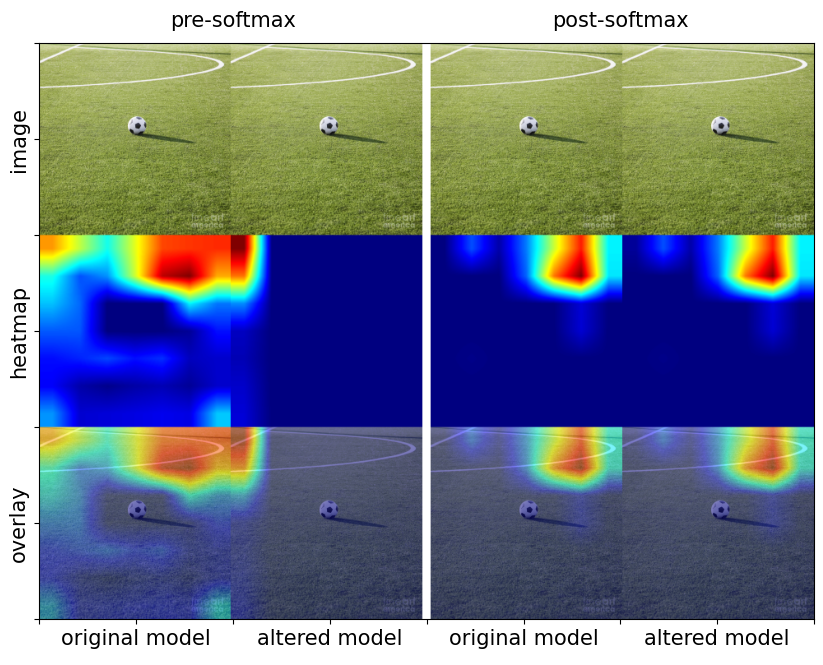}
\caption{The altered model tends to produce the same heatmap regardless
of the class assigned to the image. 
In this case Grad-CAM is used to locate a ``maze''
rather than a soccer ball in the image.  The pre-softmax version of the heatmap
on the altered model keeps highlighting the same upper left corner,
while the other heatmaps focus on the lines drawn on the grass.}\label{f:maze}
\end{figure*}

\begin{figure*}[htb]
\centering
\ \includegraphics[width=4in]{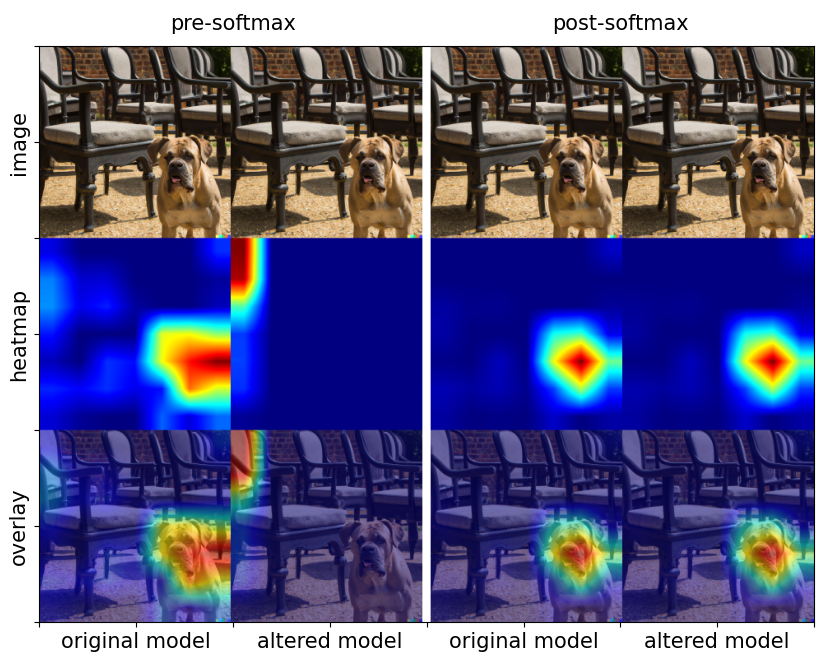}
\caption{Another example showing the heatmap computed with
pre-softmax outputs of the altered model concentrated in the upper left
corner of the image. Heatmaps computed with post-softmax outputs
remain unaltered highlighting the position of the dog.}\label{f:dog-and-chairs}
\end{figure*}

\figurename s \ref{f:soccer}--\ref{f:dog-and-chairs} show that,
for the altered model,
the heatmaps produced using pre-softmax scores are strongly distorted,
while the heatmaps produced using post-softmax scores remain unchanged.

On the other hand, since the final (post-softmax) output of the network remains unchanged, the loss function used for training 
would sit on the same local minimum for both models 
(original and modified). Further training of the models won't make
a difference since the added connection 
cannot backpropagate error. More specifically,
if $E$ is the loss function used for training, then for the modified model
we have (using multivariate chain rule):
\begin{equation}
\frac{\partial E}{\partial t} = 
\sum_{i=1}^n \frac{\partial E}{\partial y'_i}
\frac{\partial y'_i}{\partial t} = 0
\end{equation}
because $y'_i=y_i$, which does not depend on $t$, hence
$\frac{\partial y'_i}{\partial t} = \frac{\partial y_i}{\partial t} = 0$ for all $i$. 
Consequently,
the trainable parameters of both models would change in the same way, 
and if the error function $E$ is at or near a minimum for the original
model, the same would hold for the modified model.
Also, if we trained the modified VGG19 network from scratch
and with the same parameter initialization,
the final trainable parameters would be the same
as those of the original VGG19.

\section{Discussion}

We note that the main property behind the vulnerability shown here is the
possibility of altering pre-softmax scores of a classifier CNN without 
altering its post-softmax scores.
One question could be whether this vulnerability
can be exploited to deploy a malicious attack 
intended to undermine confidence in the model.
This kind of attack would be available for anybody having
access to model repositories. Since after modification the
new model would be functionally equivalent to the original one
(its outputs will not change)
it would be hard to notice that it has been modified.
Also, it is conceivable that the problem pointed out may 
manifest itself in an unintended way 
because, after training, both the original and modified model
may end up at the same local minimum
of the loss function used for training.

The phenomenon discussed may seem to have some
similarities with \emph{Clever Hans}
effects \cite{lapuschkin2016},
which also causes heatmaps to highlight wrong areas
of the input. Clever Hans effects are 
due to the ability of a classifier to
exploit spurious or artifactual correlations. For instance, in a dataset in which 
images of horses contain a watermark, the model may learn to correctly classify
the image of a horse by paying attention only to the presence of the watermark
rather than the horse.
In that case, an appropriate attribution method would  consistently highlight the area of the watermark in the images with horses, which is outside the actual 
area of interest.  However, that would not happen
because of a problem in the attribution method,
which would be correctly
revealing a problem with the model
(trained with a biased dataset).
On the contrary, the vulnerability discussed here
tells nothing about the ability of the model to extract the right information 
from the right parts of its inputs, 
it only depends on the fact that the gradients of
the pre-softmax scores may not provide the right information
to determine the impact
of the inputs on the final (post-softmax) outputs.

\section{Conclusions}

We have shown that attribution methods using pre-softmax scores are vulnerable
to a class of 
adversarial attacks that may modify the heatmaps produced without changing the model outputs.
Post-softmax outputs are not vulnerable to this kind of attack.
We have also noted that the vulnerability discussed
here is not a Clever Hans effect. Future work can be used to determine
in what extent the problem applies to a wider
class of attribution methods.

{\small
\bibliographystyle{plain}
\bibliography{mybibfile}
}

\end{document}